# CRF-based Named Entity Recognition @ICON 2013

**Arjun Das**
Computer Vision and Pattern Recognition Unit
Indian Statistical Institute
Kolkata, India
arjundas.cs@gmail.com

**Utpal Garain**
Computer Vision and Pattern Recognition Unit
Indian Statistical Institute
Kolkata, India
utpal@isical.ac.in

## Abstract

This paper describes performance of CRF based systems for Named Entity Recognition (NER) in Indian language as a part of ICON 2013 shared task. In this task we have considered a set of language independent features for all the languages. Only for English a language specific feature, i.e. capitalization, has been added. Next the use of gazetteer is explored for Bengali, Hindi and English. The gazetteers are built from Wikipedia and other sources. Test results show that the system achieves the highest F measure of 88% for English and the lowest F measure of 69% for both Tamil and Telugu. Note that for the least performing two languages no gazetteer was used. NER in Bengali and Hindi finds accuracy (F measure) of 87% and 79%, respectively.

## 1 Introduction

In the field of NLP (natural language Processing) and IR (information Retrieval) mostly in Question-answering, Named Entity Recognizer (NER) used to play a crucial role. Unlike English, development of an NER system is difficult for most of the Indian languages due to some inherent problems. For example, there exist some named entities, each of which has a valid dictionary entry. There also exists some language processing challenges, for example, there is no capitalization information, free-word ordering structure and inflectional nature of many Indic languages. Furthermore, gazetteers play an important role for developing a practical NER system. But, building well diversified gazetteers a hard and time consuming work. Hence, development of a practical NER system is still an active and challenging area of research. Though there are some research works on developing NER system for Indian languages but more research is indeed needed to develop a better insight of the problem and thereby making a practical system.

Previously IJCNLP 2008 [IJCNLP, 2008] has organized NER shared task in Indian languages (Hindi, Bengali, Punjabi, Oriya, Telugu, Urdu and English). The task involves NE's detection up to three levels. The system with first position is a hybrid [Saha et al., 2008] one, which has applied maximum entropy model for the first level NE's detection and language specific rules to identify the nested levels. In this task Ekbal et al. [Ekbal et al., 2008] have applied CRF for NE detection.

## 2 Training and Test Data

ICON has provided us the training, development and test data (in the ratio of 8:1:1) for the shared task for the Indian languages namely Hindi, Bengali, Punjabi, Telugu, Urdu and English. The details of the dataset are provided in the table 1.

| Language | No. of Sentences (Train+Dev) | No. of Sentences (Test) |
|---|---|---|
| Hindi | 4030 | 449 |
| Bengali | 3620 | 404 |
| Tamil | 4896 | 544 |
| Telugu | 1936 | 216 |
| English | 7924 | 881 |

Table 1: ICON-2013 Tool Contest Data Statistics

## 3 CRF-based NE's Detection

We have applied Conditional Random Field (CRF's) [Lafferty et al., 2001] for predicting the class lebel in the Named Entity Recognition task. In our experiment, we have used the CRF++[1], an open-source package.

We have used different combination of language independent features [Ekbal et al., 2008] and finally selected those features for which we get the highest accuracy on a development set. Following are the details of the features used in NE detection task:

- **Context Words:** The previous and next word of a particular word.
- **Word Prefix and Suffix:** We have considered word prefix and suffix from 3 character length to 5 character length for all NNP's.
- **POS and Chunk Information:** We have used POS and Chunk information of a particular word as a feature.
- **First and Last Words**: We have used the first word of a sentence and the last word i.e. (n-1)th token of the sentence as a feature.
- **Digit:** This is a binary valued feature which has been defined depending upon the presence or absence of a digit in a token.
- **Token ID:** This is a real valued feature which represents the current token id. This feature provides useful information about the position of the NE tag.
- **Associated Verb:** This feature provides the information about the nearest verb for all token. This feature is useful for differentiation between PERSON and LOACATION tag.
- **Gazetteers:** We have applied person names gazetteers for English, Bengali and Hindi. We have mined person names automatically from Wikipedia [Garain et al., 2012]. And a manually build gazetteers of Location names is applied to Bengali and English.
- **Capitalization:** For English we have applied this features. This is also a binary feature, which indicate the presence and absence of a capital letter in the first letter of the word.

## 4 Results and Error Analysis

The result of the ICON-2013 shared task is provided in the Table 2. From the results we can say that gazetteers are very helpful to boost the performance of an NER. As we have usesd gazetteers for Bengali, Hindi and English, their preformences seem to be much better with respect to Tamil and Telegu for which we did not use any gazetteers.

| Language | Precision | Recall | F-Measure |
|---|---|---|---|
| Hindi | 0.8481 | 0.7497 | 0.7959 |
| Bengali | 0.9387 | 0.8179 | 0.8741 |
| Tamil | 0.8115 | 0.6148 | 0.6996 |
| Telugu | 0.8115 | 0.6148 | 0.6996 |
| English | 0.9022 | 0.8774 | 0.8896 |

Table 2: ICON-2013 Provided Results

## 5 Conclusion

The goal of this experiment was to rapidly develop NER for Indic languages. It is experienced that CRF based approach can achieve this goal with limited accuracy. It is experienced that the role of gazetteer is equally important for Indic languages as it is for English. It is true that building diversified gazetteers is time consuming and laborious work but some tricks like use of wiki names, telephone e-directories, who's who lists, etc. may help to create gazetteers without much manual intervention. We have exercised this aspect in the present experiment and succeeded to improve the system's performance. However, more language processing might be needed to capture an NE which is basically a valid dictionary word. Use of some shallow parsing technique could be of some help for this purpose. We plan to exercise this in future.

## Acknowledgments

We would like to thanks the organizers of ICON-2013 NER Tool Contest.

---

[1]CRF++ is available at : http://crfpp.googlecode.com/svn/trunk/doc/index.html?source=navbar